\documentclass[runningheads]{llncs}
\usepackage{graphicx}
\usepackage{amssymb, amsmath, eqnarray}
\usepackage{times}
\usepackage{latexsym}
\usepackage{bm}
\usepackage{url}
\usepackage{multirow}
\usepackage{algorithm}
\usepackage[noend]{algpseudocode}

\usepackage{hyperref}
\hypersetup{colorlinks=true,
    linkcolor=blue,
    urlcolor={blue},
    filecolor={blue},
    citecolor=blue,}

\begin{document}
\newcommand\blfootnote[1]{%
\begingroup 
\renewcommand\thefootnote{}\footnote{#1}%
\addtocounter{footnote}{-1}%
\endgroup 
}
\setcounter{page}{1}
\title{Improving Multi-Head Attention with Capsule Networks}

\author{Shuhao GU\inst{1,2} \and Yang FENG\inst{1,2*}}
\authorrunning{GU. et al.}
\institute{Institute of Computing Technology, Chinese Academy of Sciences (ICT/CAS) \and
University of Chinese Academy of Sciences \\
\email{\{gushuhao17g, fengyang\}@ict.ac.cn}}

\maketitle

\begin{abstract}
  Multi-head attention advances neural machine translation by working out multiple versions of attention in different subspaces, but the neglect of semantic overlapping between subspaces increases the difficulty of translation and consequently hinders the further improvement of translation performance. In this paper, we employ capsule networks to comb the information from the multiple heads of the attention so that similar information can be clustered and unique information can be reserved. To this end, we adopt two routing mechanisms of Dynamic Routing and EM Routing, to fulfill the clustering and separating. We conducted experiments on Chinese-to-English and English-to-German translation tasks and got consistent improvements over the strong Transformer baseline. 
  \blfootnote{*Corresponding Author}

  \keywords{Neural machine translation \and Transformer \and Capsule network \and Multi-head attention}
\end{abstract}

\section{Introduction}

Neural machine translation (NMT)~\cite{bahdanau2014neural,cho2014learning,gehring2017convolutional,kalchbrenner2013recurrent,sutskever2014sequence,vaswani2017attention}
 has made great progress and drawn much attention recently. Although NMT models may have different structures for encoding and decoding, most of them employ an attention function to collect source information to translate at each time step. Multi-head attention proposed by~\cite{vaswani2017attention}
 has shown its superiority in different translation tasks and been accepted as an advanced technique to improve the existing attention functions~\cite{li2018multi,meng2018dtmt,ott2018scaling}. 

In contrast to conventional attention, the multi-head attention mechanism extends attention from one unique space to different representation subspaces. It works by first projecting the queries, keys, and values to different subspaces, then performing dot products to work out the corresponding attention in each subspace, and finally concatenating all these attentions to get the multi-head attention. This projecting process explores possible representations in different subspaces independently and hence can mitigate the all-in risk caused by one unique space.
Different attention heads may carry different features of the target sequences in different subspaces.
However, the subspaces are not always orthogonal to each other and the overlapping will lead to redundant semantic. Concatenating the attentions of different heads directly neglects the redundancy and may bring about wrong subsequent operations by treating them as different semantics, resulting in degraded performance. This also results in that only some important individual heads play consistent and often linguistically-interpretable roles and others can be pruned directly without harming the overall performance too much\cite{voita2019analyzing}.  Therefore, it is desirable to design a separate component to arrange and fuse the semantic and spatial information from different heads to help boost the translation quality of the model.

To address this problem, we propose a method to utilize capsule networks~\cite{hinton2018matrix,sabour2017dynamic} to model the relationship of the attention from different heads explicitly. 
Capsule networks provide an effective way to cluster the information of the input capsules via an iterative dynamic routing process and store the representative information of each cluster in an output capsule. Our method inserts a capsule network layer right after the multi-head attention so that the information from all the heads can be combed. We adopt two routing mechanisms, Dynamic Routing, and EM routing, for the capsule network to decide the flow of information. Then the output is feed into a fully connected feed-forward neural network. We also employed a residual connection around the input and final output layer. In our experiments, we gradually replaced the multi-head attention of the original model with ours at different positions. The experiments on the Chinese-to-English and English-to-German translation tasks show that EM routing works better than Dynamic Routing and our method with either routing mechanism can outperform the strong transformer baseline.

\section{Background}
The attention mechanism was first introduced for machine translation task by~\cite{bahdanau2014neural}. The core part of the attention mechanism is to map a sequence of $K$, the keys, to the distribution of weights $a$ by computing its relevance with $q$, the queries, which can be described as:
  \[  \mathbf a = f(\bf q, \bf K) \]
where the keys and the queries are all vectors.
In most cases, $K$ is the word embeddings or the hidden states of the model which encode the data features whereupon attention is computed. $q$ is a reference when computing the attention distribution. The attention mechanism will emphasize the input elements considered to be inherently relevant to the query.
The attention mechanism in Transformer is the so-called scaled dot product attention which uses the dot-product of the query and keys to present the relevance of the attention distribution:
    \[  {\bf a} = softmax(\frac{QK^T}{\sqrt{d_k}}) \]
where the $d_k$ is the dimensions of the keys. Then the weighted values are summed together to get the final results:
 \[   \mathbf{u} = \sum \bf a \odot V \]

Instead of performing a single attention function with a single version of a query, key, and value, multi-head attention mechanism gets $h$ different versions of queries, keys, and values with different projection functions:
    \[ Q^i, K^i, V^i = QW^Q_i, KW^K_i, VW^V_i, i\in [1,h] \]
where ${Q^i, K^i, V^i}$ are the query, key and value representations of the $i$-th head respectively. $W^Q_i, W^K_i, W^V_i$ are the transformation matrices. $h$ is the number of attention heads. 
$h$ attention functions are applied in parallel to produce the output states $\mathbf u_i$. 
Finally, the outputs are concatenated to produce the final attention:
    \[ \mathbf u = Concat(\mathbf u_1, ..., \mathbf u_h) \]

\section{Related Work}
\textbf{Attention Mechanism} Attention was first introduced in for machine translation tasks by~\cite{bahdanau2014neural} and it already has become an essential part in different architectures~\cite{gehring2017convolutional,luong2015effective,vaswani2017attention} though that they may have different forms. Many works are trying to modify the attention part for different purposes~\cite{chen2018syntax,meng2016interactive,mi2016supervised,shen2018disan,shen2018bi,tu2016modeling,yang2018modeling}. Our work is mainly related to the work which tries to improve the multi-head attention mechanism in the Transformer model.

\cite{domhan2018much} analyze different aspects of the attention part of the transformer model. It shows that the multi-head attention mechanism can only bring limited improvement compared to the 1-head model.
\cite{voita2019analyzing} evaluate the contribution made by the individual attention heads in the encoder to the overall performance of the model and then analyze the roles played by them. They find that the most important and confident heads play consistent roles while others can be pruned without harming the performance too much. We believe that because of the overlapping of the subspaces between different attention heads, the function of some attention heads can be replaced by other heads. 
\cite{li2018multi} share the same motivation with ours. They add three kinds of L2-norm regularization methods, which are the subspace, the attended positions, and the output representation, to the loss function to encourage each attention head to be different from other heads. This is a straightforward approach, but it may ignore some semantic information.
\cite{li2019information} is similar to our work, they use the routing-by-agreement algorithm, which is from the capsule network, to improve the information aggregation for multi-head attention. We did our work independently, without drawing on their work. Besides, the main structure of our model is different from theirs.
\cite{ahmed2017weighted} learn different weights for each attention head, then they sum the weighted attention heads up to get the attention results rather than just concatenating them together. \cite{shaw2018self} states that the original attention mechanisms do not explicitly model relative or absolute position information in its structure, thus they add a relative position representation in the attention function.  

\textbf{Capsule Networks in NLP} Capsule network was first introduced by~\cite{sabour2017dynamic} for the computer vision task which aims to improve the representational limitations of the CNN structure. Then~\cite{hinton2018matrix} replace the dynamic routing method with the Expectation-Maximization method to better estimate the agreement between capsules. 

There are also some researchers trying to apply the capsule network to NLP tasks. ~\cite{yang2018investigating} explored capsule networks with dynamic routing for text classification and achieved competitive results over the compared baseline methods on 4 out of 6 data sets. \cite{zhang2018attention} explored the capsule networks used for relation extraction in a multi-instance multi-label learning framework. \cite{gong2018information} designed two dynamic routing policies to aggregate the outputs of the RNN/CNN encoding layer into a final encoding layer. \cite{wang2018towards} uses an aggregation mechanism to map the source sentence into a matrix with pre-determined size and then decode the target sequence from the source representation which can ensure the whole model runs in time that is linear in the length of the sequences.

\section{The Proposed Method}
Our work is based on the multi-head attention mechanism of the Transformer model:
\begin{equation}
    \mathbf{u}_i = \mathrm{softmax}(\frac{\mathbf{Q}_i\mathbf{K}^T_i}{\sqrt{d_k}})\mathbf{V}_i; \; \; i \in [1, h]
\end{equation}
where $\mathbf{Q}_i$, $\mathbf{K}_i$ and $\mathbf{V}_i$ are computed by different versions of projection functions:
\begin{equation}
    \mathbf{Q}_i, \mathbf{K}_i, \mathbf{V}_i = \mathbf{Q}\mathbf{W}^Q_i, \mathbf{K}\mathbf{W}^K_i, \mathbf{V}\mathbf{W}^V_i, i\in [1,h]
\end{equation}
We aim to find a proper representation $\mathbf v$ based on these attention heads $\mathbf{u}$. These attention heads can be regarded as the different observations from different viewpoints on the same entity in the sequence.


Capsule network was first proposed by~\cite{sabour2017dynamic} for the computer vision tasks. 
A capsule is a group of neurons whose outputs represent different properties of the same entity. The activities of the neurons within an active capsule represent the various properties of a particular entity.
A part produces a vote by multiplying its pose matrix which is a learned transformation matrix that represents the viewpoint invariant relationship between the part and the whole. 
In the multi-head attention mechanism, different attention heads can be regarded as the different observations from different viewpoints on the same entity in the sequence. The input capsule layer represents different linguistic properties of the same input. The iterative routing process can better decide what and how much information flow to the output capsules. 
Ideally, each output capsule represents a distinct property of the input and carry all the deserved information when they are combined.

\begin{figure}[!bt]
    \centering
    \includegraphics[scale=0.175]{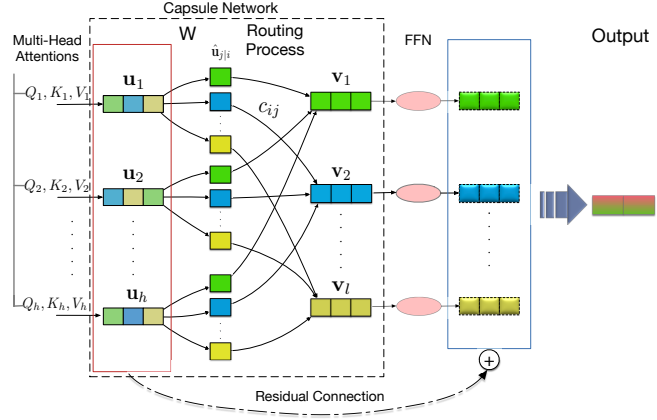}
    \caption{The architecture of our method}
    \label{fig:our}
  \end{figure}

The overall architecture is given in Figure~\ref{fig:our}. First, the capsule computes a vote by multiplying the input capsules $\mathbf u_i$ by a learned transformation matrix ${\mathbf W}_{ij}$ that represents the viewpoint invariant relationship between the part and the whole:
\begin{equation}
    {\hat {\mathbf u}}_{j|i} = {\mathbf W}_{ij}\mathbf u_i
\end{equation}
Then we compute and update the output capsules $\mathbf{v}$, the vote ${\hat {\mathbf u}}$, and the assignment probabilities $\mathbf c$ between them by a specific routing process iteratively to ensure the input to be sent to an appropriate output capsule:
\begin{equation}
    \begin{split}
        & \mathbf{v} = f({\hat {\mathbf u}}, \mathbf{c}) \\
        &\mathbf c = \mathrm{Update}({\hat {\mathbf u}}, \mathbf{v})
    \end{split}
\end{equation}

Last, the output capsules $\mathbf v$ are concatenated together and fed into a feed-forward network (FFN) which consists of two linear transformations with a ReLU activation in between:
\begin{equation}
    \mathrm{FFN}(x) = \mathrm{max}(0, \mathbf{x}\mathbf{W}_1 + \mathbf{b}_1)\mathbf{W}_2 + \mathbf{b}_2
\end{equation}
We also add a residual connection between the layer $u$ and $v$~\cite{he2016deep}. Thus the final output is:
\begin{equation}
    O = \mathbf{u} + \mathrm{FFN} (\mathbf{v}) 
\end{equation}
where
\begin{equation}
\begin{split}
    &\mathbf{u} = \mathrm{Concat}(\mathbf u_1, \dots, \mathbf u_h )  \\
    &\mathbf{v} = \mathrm{Concat}(\mathbf v_1, \dots, \mathbf v_l)
\end{split}
\end{equation}
More specifically, we have tried the Dynamic Routing and EM Routing in our method.



{\bf Dynamic Routing} In this method, 
 we sum up all these weighted vote vectors to get the origin output capsule vectors:
\begin{equation}
\begin{split}
    &{\mathbf s}_j = \sum_i c_{ij}{\hat {\mathbf u}}_{j|i} \\
\end{split}
\end{equation}
where
\begin{equation}
 c_{ij} = \frac{\exp(b_{ij})}{\sum_k \exp(b_{ik})}
\end{equation}
 the $c_{ij}$ are determined by computing the "routing softmax" of the initial logits $b_{ij}$ which are initialized to zero.

Next, the origin output capsule vectors ${\mathbf s}_j$ is applied with a squashing function to bring non-linearity to the whole model:
\begin{equation}
    \mathbf{v}_j = \frac{||{\mathbf s}_j||^2}{1+||{\mathbf s}_j||^2} \frac{{\mathbf s}_j}{||{\mathbf s}_j||}
\end{equation}

The initial coupling coefficients $b_{ij}$ are iteratively refined by measuring the agreement between the current output $\mathbf{v}_j$ by the dot-product of the input capsules and each output capsule:
\begin{equation}
    b_{ij} \leftarrow b_{ij} + {\hat {\mathbf u}}_{j|i} \cdot \mathbf{v}_j
\end{equation}



\begin{table*}[]
    \centering
    \renewcommand\arraystretch{1.3}
    \begin{tabular}{l||c|c c c c c||c}
    \hline
    \textbf{Zh$\rightarrow$En} & \# Para.  & MT03 & MT04 & MT05 & MT06 & MT08 & AVE. \\
    \hline 
    \hline
    \cite{wang2018neural}\ \ \ \  &- & 46.60 & 47.73 & 45.35 & 43.97 & - & -\\
    \cite{zhang2018regularizing} &-& 48.28 & - & 46.24 & 46.14 & 38.07 & - \\
    \hline
    Transformer-Base &84.6M & 46.70 & 47.68 & 47.04 & 46.16 & 37.19 & 44.95 \\
    + Dynamic Routing &92.2M& 47.60 & 48.04 & 47.30 & 46.56 & 37.80 & 45.46 (+0.51) \\
    + EM Routing &91.7M& 47.62 & 48.07 & 47.97* & 46.80 & 38.01* & 45.70 (+0.75) \\
    \hline
    \end{tabular}
    \caption{Case-insensitive BLEU scores for Zh$\rightarrow$En translation. ``\# Para'' denotes the number of parameters. ``*'' is used to indicate the improvement is statistically significant
    with $\rho < 0.05$~\cite{collins2005clause}.}
    \label{tab:main_results}
\end{table*}

\textbf{EM Routing} 
In this method, each capsule becomes a combination of a $n\times n$ pose matrix, $M_i$, and an activation probability, $\alpha_i$. Each output capsule corresponds to a Gaussian distribution and the pose matrix of each active capsule in the lower-layer corresponds to a data-point. The iterative routing process is a version of the Expectation-Maximization procedure which iteratively adjusts the means, variances and activation probabilities of the output capsules and the assignment probabilities between the two layers. The whole procedure can be divided into two parts:

\textbf{M-Step} Keep the assignment probabilities between the two layers fixed and compute the mean $\mu_j$ and variance $\sigma_j$ of the output capsules:
\begin{equation}
    \begin{split}
         \mathbf{v}_j & = \frac{\sum_i c_{ij} {\hat {\mathbf u}}_{j|i}}{\sum_i c_{ij}} \\
         (\sigma_j)^2 & = \frac{\sum_i c_{ij}({\hat {\mathbf u}}_{j|i}-\mu_j)^2}{\sum_i c_{ij}}
    \end{split}
\end{equation}
Then we compute the incremental cost and the activation probability:
\begin{equation}
    \small
    \begin{split}
       & cost_j  = (\log(\sigma_j)+\frac{1+\ln{2\pi}}{2})\sum_i c_{ij} \\
       & \alpha_j  = \mathrm{logistic}(\lambda(\beta_\alpha - \beta_\mu \sum_i c_{ij} - \sum_h cost_j^h))
    \end{split}
\end{equation}
where $\sum_i c_{ij}$ is the amount of data assigned to $j$. We learn $\beta_\alpha$ and $\beta_\mu$ discriminatively and set a fixed schedule for $\lambda$ as a hyper-parameter.

\textbf{E-Step} Keeping the Gaussian distributions of the output capsules fixed, we need to calculate the incremental cost of explaining a whole data-point $i$ by using an active capsule $j$:
\begin{equation}
    p_j = \frac{1}{\sqrt{2\pi(\sigma_j)^2}}\exp{-\frac{({\hat {\mathbf u}}_{j|i}-\mathbf{v}_{j})^2}{2(\sigma_j)^2}}
\end{equation}
and then adjust the assignment probabilities based on this:
\begin{equation}
    c_{ij} = \frac{\alpha_j p_j}{\sum_k \alpha_k p_k}
\end{equation}
Please refer to~\cite{hinton2018matrix} for more details. The output capsules are then reshaped into vectors and also fed into the feed-forward network. 
We just make use of the higher capsule as the representation of the attention results and abandon its activation probabilities to take advantage of the information aggregation way of the capsule network.

\section{Experiments}
We evaluated our method on the NIST Chinese$\rightarrow$English (Zh$\rightarrow$En) and WMT14 
\\English$\rightarrow$German (En$\rightarrow$De) translation tasks.

\subsection{Setup} 
\textbf{Chinese$\rightarrow$English} The training data consists of about 1.25M sentence pairs from LDC corpora with 27.9M Chinese words and 34.5M English words respectively \footnote{The corpora include LDC2002E18, LDC2003E07, LDC2003E14, Hansards portion of LDC2004T07, LDC2004T08 and LDC2005T06.}. We used NIST 02 data set as the development set and NIST 03, 04, 05, 06, 08 sets are used as the test sets. We tokenized and lowercased the English sentences using the Moses scripts\footnote{http://www.statmt.org/moses/}. For the Chinese data, we performed word segmentation using the Stanford Segmentor\footnote{https://nlp.stanford.edu/}. Besides 30K merging operations were performed to learn byte-pair encoding(BPE)~\cite{sennrich2015neural} on both sides.

\textbf{English$\rightarrow$German} For this task, we used the WMT14 corpora pre-processed and released by Google
\footnote{https://drive.google.com/uc?export=download\&id=0B\_bZck-ksdkpM25jRUN2X2UxMm8}
which consists of about 4.5M sentences pairs with 118M English words and 111M German words. We chose the newstest2013 as our development set and newsset2014 as our test set.

We evaluate the proposed approaches on the Transformer model and implement it on the top of an open-source toolkit - Fairseq-py~\cite{Edunov2017fairseq}. We follow~\cite{vaswani2017attention} to set the configurations and have reproduced their reported results on the En$\rightarrow$De task with both of the Base and Big model. All the models were trained on a single server with eight NVIDIA TITAN Xp GPUs where each was allocated with a batch size of 4096 tokens. 
The routing iterations are set to 3 and the number of output capsules is set to equal to the number of input capsules if there is no other statement.

During decoding, we set beam size to 4, and length penalty $\alpha$=0.6. Other training parameters were the same as the default configuration of the Transformer model.

\begin{table}[tbp]
    \centering
    \renewcommand\arraystretch{1.1}
    \begin{tabular}{l||c c}
       \textbf{En$\rightarrow$De} & \# Para. & BLEU  \\
        \hline
        \hline
         Transformer-Base& 60.9M & 27.34 \\
         + Dynamic Routing &62.0M  & 27.67 \\
         + EM Routing &61.6M  & 27.77  \\
         \hline
         Transformer-Big &209.9M & 28.43 \\
         +Dynamic Routing & 216.2M &28.65\\
         +EM Routing & 214.2M &28.71 \\
         \hline
         
    \end{tabular}
    \caption{Case-sensitive BLEU scores for En$\rightarrow$De translation.}
    \label{tab:res_ende}
\end{table}

\subsection{Main Results}
We reported the case-insensitive and case-sensitive 4-gram NIST BLEU score~\cite{papineni2002bleu} on the Zh$\rightarrow$En and En$\rightarrow$De tasks, respectively. During the experiments, we found that our proposed method achieved the best performance when we only insert the capsule network after the last multi-head attention sub-layer in the decoder and in the attention layer between source and target. We will analyze this phenomenon in detail in the next subsection.


The Zh$\rightarrow$En results of the Transformer-Base model are shown in the Table~\ref{tab:main_results}. Both of our models (Row 4,5) with the proposed capsule network attention mechanism 
can not only outperform the vanilla Transformer(Row 3) but also achieve a competitive performance compared to the state-of-the-art systems(Row 1,2, we use the results from the related paper directly), indicating the necessity and effectiveness of the proposed method. It shows that our method can get the information well combed and preserve all the deserved information.

Among them, the '+EM Routing' method is slightly better than the '+Dynamic Routing' method by 0.24 which because of better estimating the agreement during the routing. Besides, it requires fewer parameters and runs much faster. Considering the training speed and performance, the '+EM Routing' method is used as the default multi-head aggregation method in subsequent analysis experiments.


The En$\rightarrow$De results are shown in the Table~\ref{tab:res_ende}. In this experiment, we have applied our proposed methods both on the Base and Big model. The results show that our model can still outperform the baseline model, indicating the universality of the proposed approach.

\begin{table}[tbp]
    \centering
    \renewcommand\arraystretch{1.1}
    \begin{tabular}{l|c}
         \# Model Varations & NIST 04 \\
         \hline
         \hline
         Transformer-Base & 47.68 \\
         $\mathrm{Enc}_\mathrm{1}$ & 47.25 (-$\mathtt{0.43}$) \\
         $\mathrm{Enc_1}, \mathrm{Enc_2}$ & 47.19 (-$\mathtt{0.49}$) \\
         $\mathrm{Enc_5, Enc_6}$ & 47.64 (-$\mathtt{0.04}$) \\
         $\mathrm{ED_1}$ & 47.38 (-$\mathtt{0.30}$)\\
         $\mathrm{ED_6}$ & 47.98 (+$\mathtt{0.30}$) \\
         $\mathrm{ED_5, ED_6}$ & 47.43 (+$\mathtt{0.11}$) \\
         $\mathrm{Dec_1}$ & 47.45 (-$\mathtt{0.23}$) \\
         $\mathrm{Dec_6}$ & 48.03 (+$\mathtt{0.35}$) \\
         $\mathrm{Dec_5, Dec_6}$ & 47.83 (+$\mathtt{0.15}$) \\
         $\mathrm{ED_6, Dec_6}$ & 48.07 (+$\mathtt{0.39}$) \\
         $\mathrm{ED_5, ED_6, Dec_5, Dec_6}$ & 47.99 (+$\mathtt{0.31}$) \\
         \hline

    \end{tabular}
    \caption{Case-insensitive BLEU scores for different ways of integrating capsule networks. $\mathrm{Enc_i}$, $\mathrm{Dec_i}$, $\mathrm{ED_i}$ mean the capsule network is inserted after $\mathrm{i}$-th multi-head attention sub-layer in the encoder, in the decoder and in the attention layer between source and target, respectively. For example, $\mathrm{Dec_6}$ means the capsule network is inserted between the multi-head attention and the FFN in the 6th layer of the decoder.}
    \label{tab:res_2}
\end{table}

\begin{figure}[!bt]
    \centering
    \includegraphics[scale=0.5]{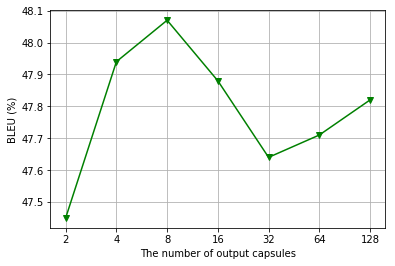}
    \caption{Impact of the number of output capsules. }
    \label{fig:number}
  \end{figure}

\subsection{Impact of Different Ways to Integrate Capsule Networks}
The Transformer model consists of three kinds of attention, including encoder self-attention, encoder-decoder attention and decoder self-attention at every sublayer of the encoder and decoder. We gradually insert the capsule network in different places and measured the BLEU scores on the NIST 04 test set based on the "+EM Routing" Transformer-Base model. The results are shown in the Table~\ref{tab:res_2}. It shows that not all of the changes are positive to the results. 

First, any changes to the encoder self-attention, no matter to the top sublayer($\mathrm{Enc_5}, \mathrm{Enc_6}$) or the bottom sublayer($\mathrm{Enc_1}, \mathrm{Enc_2}$) of the encoder is harmful to the performance of the whole model. One possible reason for this may be that the routing part should be close to the supervisory signals to be well trained. Without its help, the capsule network only extracts internal features regardless of whether these features are helpful to the translation quality. Another reason for this may be that although we add a residual connection between the input capsule layer and the output capsule layer to ensure preserve all the information, we don't add the reconstruction procedure of the origin work~\cite{sabour2017dynamic} which may make the output leave out some information inevitable.

Then, the changes to the bottom sublayer of the encoder-decoder attention($\mathrm{ED_1}$) and the bottom sublayer of the decoder self-attention($\mathrm{Dec_1}$)  also degrade the performance, which is also far from the supervisory signals.

Last, the changes to the top sublayer of the encoder-decoder attention($\mathrm{ED_5, ED_6}$) and the top sublayer of the decoder self-attention($\mathrm{Dec_5, Dec_6}$) are beneficial for the final results because that they are more close to the output layer which supports our hypothesis above.


\subsection{Impact of the Number of Output Capsules} The number of output capsules $l$ is a key parameter of our model. We assumed that the capsule network can capture and extract high-level semantic information. But it is not obvious how much high-level information and what kind of information can be aggregated. Therefore we varied the number of the output capsules and also measured the BLEU scores on the NIST 04 test set based on the "+EM Routing" Transformer-Base model. The results are shown in Figure~\ref{fig:number}. It should be mentioned that the dimension of each output capsule is set to $d/l$ to keep the final output be consistent with the hidden layer. The results show that our proposed method achieves the best performance when the number of output capsules is equal to the number of input capsules. 

\begin{figure}[!bt]
    \centering
    \includegraphics[scale=0.25]{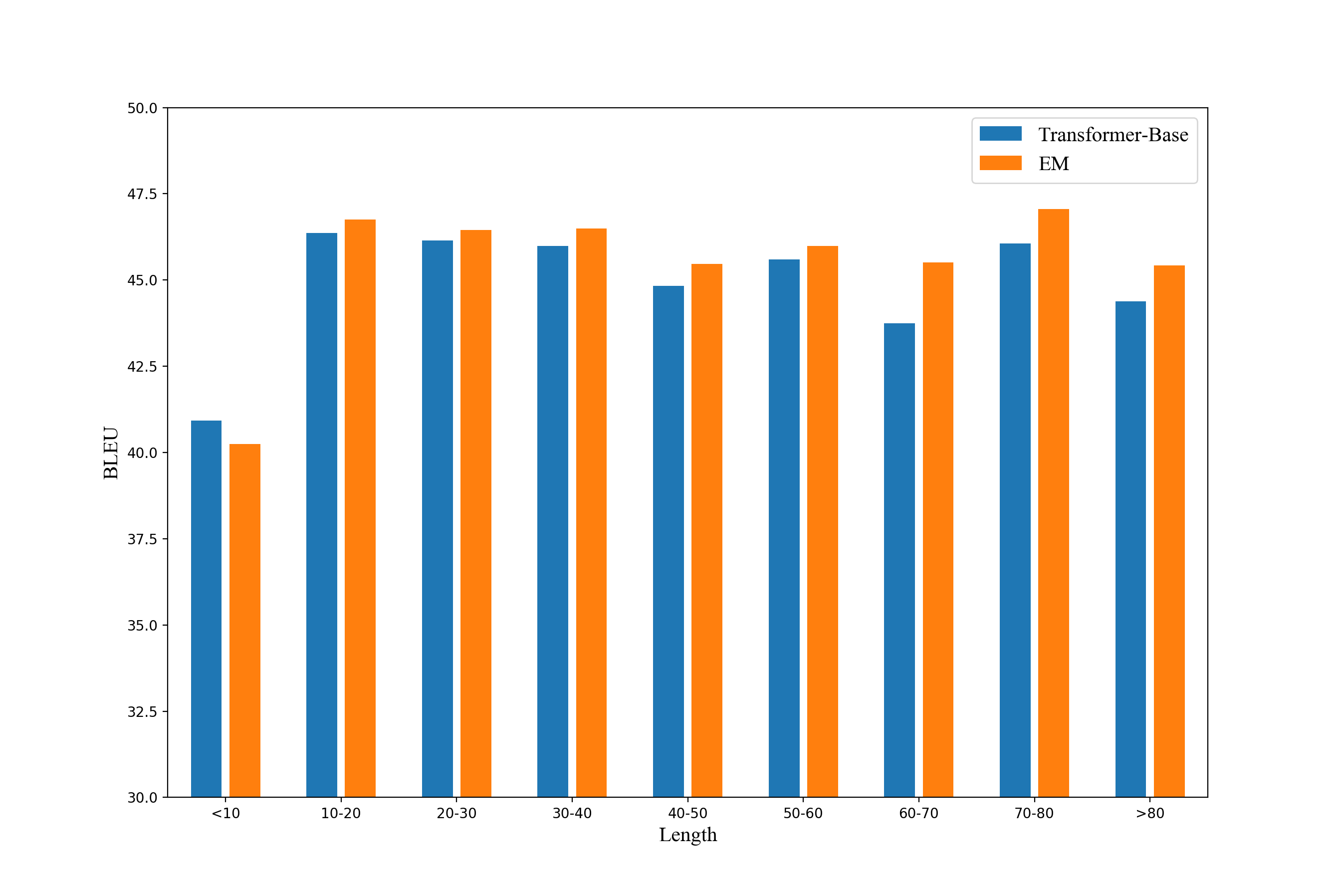}
    \caption{Effect of source sentence lengths.}
    \label{fig:length}
\end{figure}

\subsection{Effect of Source Sentence Length}
We also evaluated the performance of the best version of our proposed method '+EM Routing' and the baseline on the combined NIST 03-08 test set with different source sentence lengths. The results are shown in Figure~\ref{fig:length}.In the bins holding sentences no longer than 60, the BLEU scores of the two systems are close to each other. When the sentence length surpasses 60, our method shows its superiority over the Transformer base model. As the sentence length grows, the difference becomes increasingly large. That is because our method provides an effective way to cluster the information of the multi-head results so that it can get information well aggregated especially when the sentence lengths increase and handle more information.

\section{Conclusion}
In this work, we argue that the neglect of semantic overlapping between subspaces of the different attention heads increases the difficulty of translation. 
We adopt the Dynamic Routing and EM Routing and evaluated our methods on popular translation tasks of different language pairs and the results showed that our method can outperform the strong baselines. The extensive analysis further suggests that it can help to improve the translation performance only when we set the capsule part close to the supervisory signals. 

\section*{Acknowledgements}
This work was supported by the National Natural Science Foundation of China (NSFC) under the project NO.61876174, NO.61662077 and NO.61472428.



\end{document}